# Language-free Experience at Expo 2025 Osaka

**Michael Paul, Kenji Imamura, Xiaolin Wang, Shohei Higashiyama, Masao Utiyama**
National Institute of Information and Communications Technology (NICT), Japan
{michael.paul, kenji.imamura, xiaolin.wang,
shohei.higashiyama, mutiyama}@nict.go.jp

## Abstract

In line with the Global Communication Plan 2025, we have pursued the development of multilingual translation technologies to realize a language-barrier-free experience at Expo 2025 Osaka. Our work includes the advancement of simultaneous interpretation systems emphasizing high translation quality and low latency. Key achievements include chunk-based input segmentation, context-aware translation, and multi-engine machine translation technologies. Through demonstration deployments and collaboration with private companies, our technologies have led to real-world applications, with several services and systems showcased at Expo 2025 Osaka.

## 1 Introduction

Simultaneous interpretation is a task that converts a speaker's speech into another language in almost real time while listening to the speaker's speech. The interpreter is required to have a high degree of focus and agility, as well as deep knowledge of both languages. However, the burden of interpreters is high, and problems such as time constraints, differences between languages, and barriers of specialized knowledge and terminology arise, and they are not easily available in daily life situations.

Due to the emergence of Neutral Machine Translation (NMT) (Vaswani et al., 2017), the quality and practicality of machine translation tools improved significantly in recent years. It has become possible to translate long sentences, such as document translation with high accuracy. As for spoken languages, practical applications have been developed that allow you to start translation even in the middle of a conversation, like simultaneous interpreters.

This paper introduces the multilingual simultaneous interpretation system being developed at the National Institute of Information and Communications Technology (NICT). The system translates spoken language like lectures with high quality and low delay, without interrupting the speech. It covers the 15 languages listed in Table 1.

| Code | Language | Code | Language |
|---|---|---|---|
| ja | Japanese | mn | Mongolian |
| en | English | my | Myanmar |
| es | Spanish | ne | Nepali |
| fp | Filipino | pt | Brazilian |
| fr | French | (BR) | Portuguese |
| id | Indonesian | th | Thai |
| km | Khmer | vi | Vietnamese |
| ko | Korean | zh | Chinese |

**Table 1:** Supported Languages

To reduce the delay in translation (latency), it is necessary to translate while the speaker is still speaking. However, in general, the quality of the translation will deteriorate if the translation is performed before sufficient context has been heard. To reduce latency without sacrificing translation quality, NICT's system translates using chunks. A chunk imitates the unit in which an interpreter reads an utterance when performing sequential interpretation, and is a unit shorter than a sentence, such as a few words

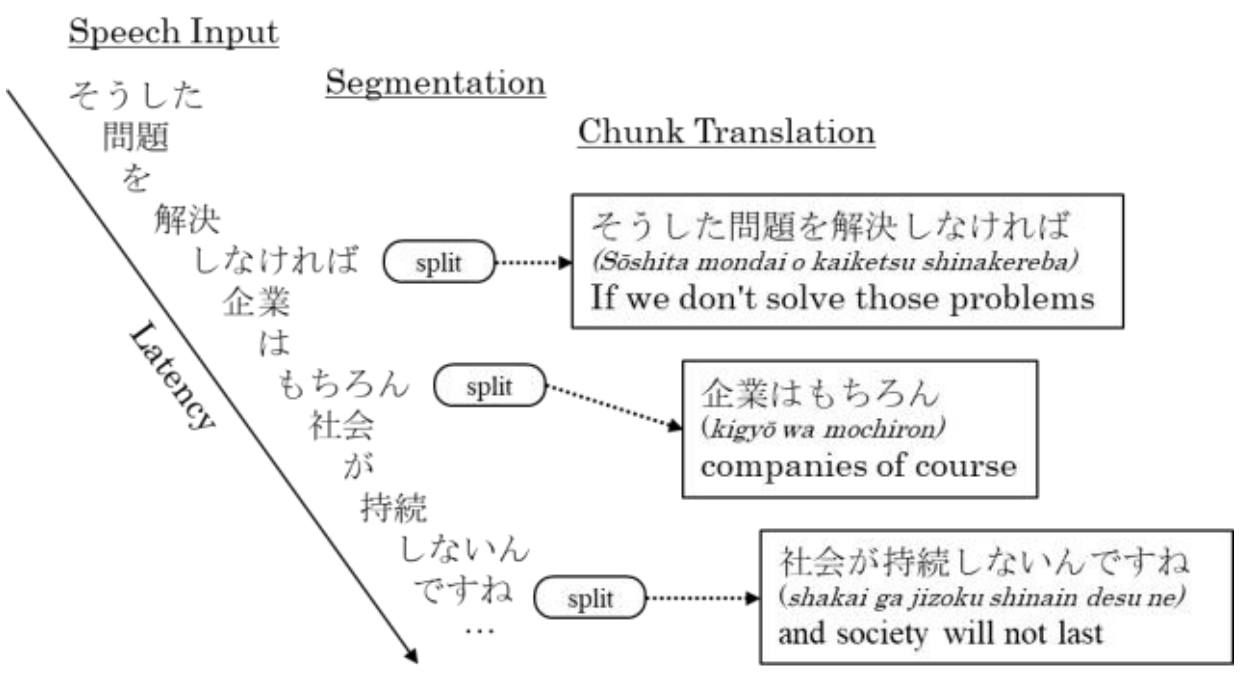

**Figure 1:** Chunk-based simultaneous interpretation

Figure 1 gives an example of how speech input provided by an automatic speech recognition engine is automatically segmented into chunks, whereas all chunks are translated sequentially to obtain a simultaneous interpretation of the speech input.

Figure 2 shows the structure of the simultaneous interpretation system where speech recognition results are segmented online into meaningful chunks and translated using models learned from multiple conversation corpora.

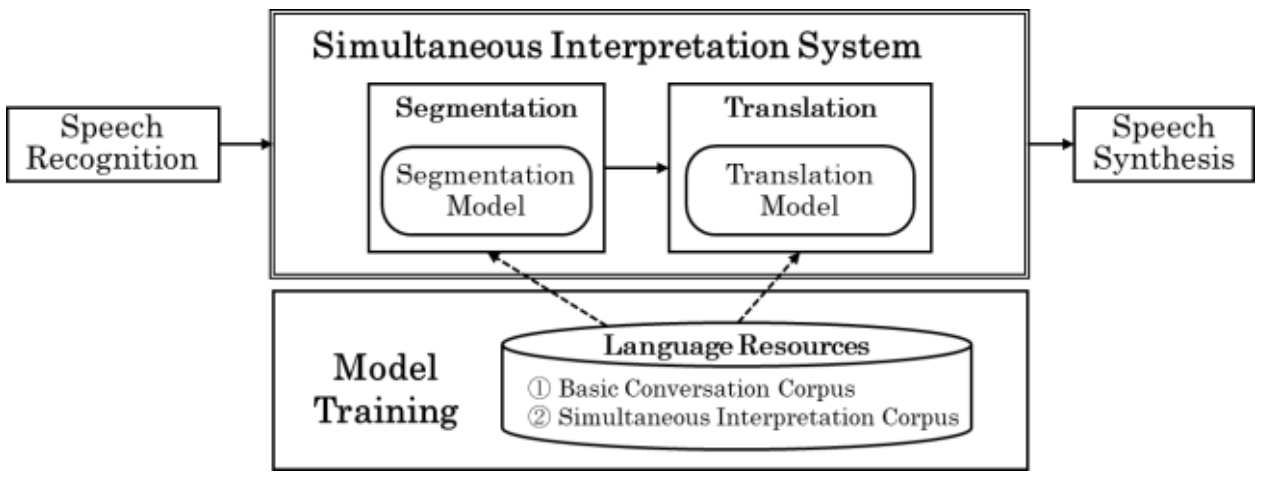

**Figure 2:** Simultaneous interpretation system

The language resources are used to learn multi-lingual input segmentation and translation models, as well as the context processing required to understand the conversational flow and to generate high-quality and easy-to-understand translations. Finally, we summarize evaluation results of translation latency and translation quality proving the practicability of automatic simultaneous interpretation systems.

## 2 Language Resources

The language resources were created within the Global Communication Plan initiative launched by the Ministry of Internal Affairs and Communications (Japan) in 2014 aiming at the development of multilingual speech and text translation technologies that enable seamless communication across languages in various public and private sectors.

### 2.1 Basic Conversation Corpus

To ensure basic translation quality, we developed a multilingual conversation corpus focusing on simulated dialogs (Imamura and Sumita, 2018). It was designed to cover domains such as healthcare, disaster prevention, shopping, tourism, and others. An example of such a simulated dialog in the medical field is shown in Table 2. The multilingualization of simulated dialogs was achieved by translating Japanese into foreign languages by humans.

【Domain】Medical Care / Responding to Emergencies
【Speaker】Foreigner (F), Japanese (J)

| | Japanese | English |
|---|---|---|
| F | このあたりに病院はありませんか？ | Is there a hospital nearby? |
| J | あのコンビニを右に曲がったところに内科の医院があります。 | There is an internal medicine hospital after you turn right at that convenience store. |
| J | 具合が悪いんですか？ | Aren’t you feeling well? |
| F | めまいがします。 | I feel dizzy. |
| J | 日本語は得意ではないんですか？ | Aren’t you good at Japanese? |
| F | 得意ではありません。 | I am not good at it. |
| J | あそこの医院だと英語は通じないかもしれません。 | It may be impossible to communicate in English in the hospital there. |

**Table 2:** Japanese dialog with English translation

This corpus contains spoken language, so it can be used to translate lectures, etc. However, these are not real utterances, but written by humans, and thus do not contain fillers and re-phrasing, which are characteristics of spoken language. The corpus contains additional information such as speakers and domains, which can be used to adapt the translation models to specific conversation scenarios.

### 2.2 Simultaneous Interpretation Corpus

To enable the training of translation and input segmentation models suitable for simultaneous interpretation, we constructed a multilingual simultaneous interpretation corpus centered on Japanese. Due to the heavy workload and high costs to record and transcribe human speech and its human simultaneous interpretation, we adopted the following method to construct the simultaneous interpretation corpus.

- Prepare the original text to be interpreted. The original text includes transcripts of real speech and conversations. Real speech resources include lectures on various fields and topics.
- A human interpreter divides the original text into semantic chunks that are translated one at a time, while assuming a situation in which the original text is read aloud. The simultaneous interpretation is then obtained by the concatenation of sequential chunk translations
- The original Japanese transcriptions were translated into the 14 languages listed in Table 1 and for 7 languages (en, fr, ko, th, fp, vi, zh) original transcriptions were prepared accordingly and translated into Japanese.

Using this method, a simultaneous interpretation corpus was constructed for each of the above-mentioned languages and used to train input segmentation and simultaneous interpretation models. As an example of the simultaneous interpretation corpus, the original text, simultaneous interpretation, and standard sentence translation of a Japanese English business conversation is shown in Table 3.

| | |
|---|---|
| Original Text | 先日ご紹介した商品同様、/ 20 年未満の積立期間だと、/ 途中解約した場合、/ 戻ってくるお金は積立金の 0.8 倍になります。 |
| Simultaneous Interpretation | Same as the product I introduced the other day, / if the funding term is less than 20 years, / and if you cancel it before full term, / 80% of your funds will be returned. |
| Original Text Translation | If you cancel your plan before reaching the 20 year saving stage, you’ll be reimbursed only 80% of your saving, same as the product I showed you the other day. |

**Table 3:** Simultaneous interpretation corpus

The semantic chunks divided by the interpreter are marked by a slash. The translation of the original text begins with "If you cancel your plan", which is equivalent to the third chunk of the simultaneous interpretation "If you cancel it before full term".

Therefore, a standard translation approach would have to delay the processing of the first chunk until the second and third chunk are translated to reproduce the word order of the original text translation.

The simultaneous interpretation approach, however, translates each chunk of the original text sequentially, thus minimizing translation latency.

Details on the construction of the initial Japanese English simultaneous interpretation corpus are described in (Higashiyama et al., 2023).

## 3 Simultaneous Interpretation

In this section, we explained in detail how the speech transcription input is segmented automatically into semantic chunks, how context processing for simultaneous interpretation is carried out, and how multiple translation engines can be exploited to improve translation quality.

### 3.1 Input Segmentation

In an automated simultaneous interpretation system, it is usually necessary to integrate two basic natural language processing techniques, Automated Speech Recognition (ASR) and Machine Translation (MT). However, chunk or sentence segmentations are not provided in transcriptions generated by an ASR system. On the other hand, for machine translation to work accurately, chunk-wise or sentence-wise input needs to be provided to the MT system.

NICT's simultaneous interpretation system uses an online segmentation technique that naturally bridges the gap between ASR and MT components by dividing a stream of input words into semantic chunks that can be translated by the MT engine in real time. The difficulty of the online input segmentation problem is mainly due to two factors (Wang et al., 2019).

Firstly, the ASR engine cannot reliably segment the speech input into unpunctuated word sequences that are meaningful enough to be translated by the MT engine. To address this issue, we used a recursive neural network architecture commonly employed in language models. This model operates online, receiving a word as input at each step and returning boundary information for sentence or chunk segmentation. Additionally, due to the mechanism of the network, it retains the input received up to the current time.

Secondly, whether there is a sentence boundary at the current position depends on the following words, and the optimal number of words to wait for varies depending on the context. Therefore, we trained the segmentation model to simultaneously decide whether there was a boundary or not for up to N successive words, whereby the factor N depends on the input language and is automatically optimized using the development sets of the interpretation corpus reserved for parameter tuning (Wang et al., 2019).

In the example shown in Figure 3, the first 'Yes' signal is output after the word 'find' is input. This corresponds to two words earlier, indicating that the chunk ended after the word 'tea'.

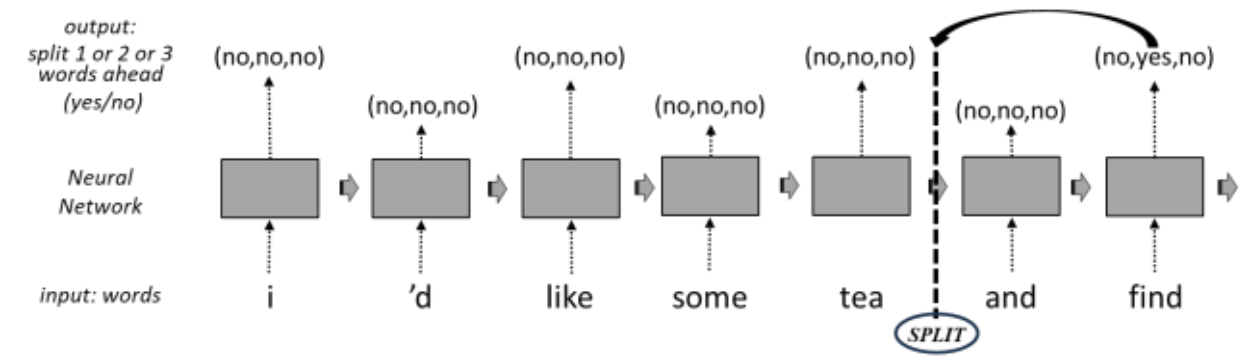

**Figure 3:** Segmentation detection considering a variable number of subsequent words

NICT's simultaneous interpretation system uses two types of input segmentation models, a chunk-level segmentation model and a sentence-level segmentation model, which are used for low-latency chunk translation and high-accuracy retranslation of the input chunk sequence at the end of each detected sentence, respectively.

### 3.2 Context Processing

Context processing in machine translation is a technique that aims for more natural translation by understanding the relationships between words and the flow of sentences, as well as grasping the conversational setting (who is talking about what topic) and the content of previous utterances in the conversation.

NMT approaches enable the incorporation of contextual information when learning translation models from bilingual sentences. By attaching context tags to the training resources, the semantic interpretation of the same input can be switched, or different output variations can be generated.

NICT's simultaneous interpretation system added the context tags listed in Table 4 to the training resources, and domain adaptation was performed on the standard translation model.

| Category | Context Tag | Language |
|---|---|---|
| speaker | Japanese, foreigner | all |
| scene | business, disaster, education, medical, municipality, shopping, sightseeing, sports, transportation, others | all |
| subject | watashi ("I"), anata ("you") | Japanese |
| gender | female, male | Thai |

**Table 4:** Context tags for dialog-aware translation models

By adding conversational context (such as speaker tags and scene tags) to the beginning of the input text, it became possible to produce translations that align with the given dialog, as shown in the Japanese English translation sample in Table 5.

**(Input)**
さらに昨日は咳が出て止まらないようでした
**(Reference Translation)**
And last night, **he kept coughing**.

| Context | Translation |
|---|---|
| none | In addition, yesterday **I had a cough** that did not seem to stop. |
| foreigner medical | In addition, **he coughed** and couldn't stop it yesterday. |

**Table 5:** Dialog-aware MT output

Moreover, regarding contextual processing, it is important to grasp the content of the previous conversation in addition to the conversational setting. Sentence-by-sentence translation has long been the mainstream approach, but recent neural machine translation has made it possible to utilize context that spans multiple sentences (Tiedemann and Scherrer, 2017).

NICT's simultaneous interpretation system memorizes past conversations and their translation results, and when translating the next utterance, this information is provided as extra-sentential context. This helps supplement appropriate expressions by referring to the flow of the conversation, especially when the meaning of the input sentence alone is ambiguous.

### 3.3 Multi-Engine Translation

Multi-engine translation technology enables more natural and high-quality translations by using multiple translation engines in parallel and selecting the most appropriate engine online.

Table 7 lists the translation engines and corresponding language pairs that can be used in NICT's simultaneous interpretation system.

| Type | Engine | Translation Models |
|---|---|---|
| NMT | General (GPMT) | Multilingual translation model supporting 54 language pairs |
| NMT | Adaptation (TSEG) | Domain adaptation training model based on the GPMT model |
| NMT | Universal (UNIV) | Translation of 32×31 languages with a single model |
| LLM | Universal (RWKV) | Multilingual translation model with contextual understanding |

**Table 7:** Translation Engines

The general-purpose machine translation (GPMT) models are high-quality NMT models for the 14 language-pairs (from/to Japanese) and the 13 language-pairs (from/to English) in Table 1, trained on the Basic Conversation Corpus.

The translation models of the adaptive MT engine (TSEG) are trained by adding the simultaneous interpretation corpus to the learning resources of the GPMT translation model and including the context tag information described in the previous section.

The translation model of the universal MT engine (UNIV) is a single translation model that is trained by combining the learning resources of all GPMT language pairs. The input and output languages can be switched at runtime.

In addition to the NMT technology, machine translation methods based on large-scale language models (LLM) have also attracted much attention (Higashiyama, 2024). The LLM approach features excellent general language understanding and contextual processing, enabling the generation of natural and flexible translations. The simultaneous interpretation system uses RWKV's LLM model.

To select the most appropriate translation from multiple translation engines, the system uses back-translation which is a technique that uses the reverse-direction translation model of each MT engine to translate the output utterance back into the original language. The similarity of the original sentences and the back translation results is calculated by vectorizing the word sequences and computing their cosine similarity.

Table 8 shows the translation results of the input, the back-translation results of the translation, and its similarity score of the respective MT engines. For this example, the translation result of the RWKV engine is selected.

## 4 Usability Design and Evaluation

The automation of simultaneous interpretation reduces problems such as cross-language differences and professional knowledge and terminology barriers that burden interpreters, but the challenge of time constraints remains. The slower the response time (time from speaking to translation) of an automatic simultaneous interpretation system, the slower the flow of conversation, and the lower the usability of the automatic simultaneous interpretation system.

### 4.1 Realization of Chunk Translation

To realize a simultaneous interpretation system that provides real-time responses, NICT's simultaneous interpretation system combined sentence-level translation with chunk-level translation processing. A "chunk" is a unit of short semantic meaning that can be translated, which is inferred from the original text during simultaneous interpretation.

More specifically, based on the sentence-level segmentation model introduced in Section 3.1, we used the chunk information of the simultaneous interpretation corpus in Section 2.2 to train the segmentation model for each language to segment chunks shorter than a sentence. As a result, we achieved automatic simultaneous interpretation with low latency and practical accuracy. However, since the amount of information in a chunk is less than that in a full sentence, the translation quality of the chunk-based approach is lower than that of the sentence-level approach, There-

**(Input)**
逆にマグニチュードの大きい地震でも震源から遠いと震度は小さくなるということですね
**(Reference Translation）**
On the contrary, even in a large magnitude earthquake, the seismic intensity can be smaller in the location which is far away from the focus.

| Engine | Translation | Back Translation | Similarity |
|---|---|---|---|
| GPMT | On the other hand, even if the magnitude of an earthquake is large, if it is far from the epicenter, the seismic intensity becomes smaller. | 逆に、地震のマグニチュードが大きくても、震源地から離れていれば震度は小さくなります。 | 0.4930 |
| ADAPT | On the other hand, even if the magnitude is large, the seismic intensity becomes small if it is far from the epicenter. | 逆にマグニチュードが大きくても、震源地から遠いと震度は小さくなるのですね。 | 0.6789 |
| UNIV | On the contrary, even for an earthquake with a large magnitude, the seismic intensity becomes smaller if it is far from the epicenter. | 逆に、マグニチュードの大きい地震でも、震源から離れれば震度は小さくなります。 | 0.6781 |
| RWKV | **Conversely, even if the magnitude of an earthquake is large, if it is far from the epicenter, the seismic intensity will be small.** | **逆に、マグニチュードの大きい地震でも震源から遠いと震度は小さくなるということですね。** | **0.8907** |

**Table 8:** Selection of multiple engine translation results

fore, it is necessary to determine whether the input needs to be corrected or not when the input is automatically translated using chunks.

In NICT's simultaneous interpretation system, chunk and sentence translation processing is parallelized. When the chunk boundaries align with sentence boundaries, sentence-level retranslation is performed by replacing the sequence of chunk-level translation results with the full sentence translation output.

### 4.2 Simultaneous Interpretation Evaluation

Chunk-based simultaneous interpretation must always wait for one chunk to be input before starting translation, so the shorter the chunk, the less the delay. Therefore, we evaluated the length of chunks and sentences after segmentation. We simulated the result of speech recognition by concatenating multiple input utterances and compared the segment length when processed by the "sentence segmentation engine" and the "chunk segmentation engine", respectively. The results confirmed that the chunk segmentation engines reduced chunk length by 20% to 60%.

Figure 4 shows the average number of words in (a) the original utterance (b) the sentence-level segmentation and (c) the chunk-level segmentation of the simultaneous interpretation corpus test sets.

For Japanese, an average of 23.6 input word utterances where split into an average of 12.8 and 7.8 words by the sentence-level and chunk-level segmentation model, resulting in a 39% reduction in segment length for the chunk-based segmentation output. Similar investigations for other languages showed that chunk segmentation consistently resulted in significant reductions in length across all languages. Therefore, the time interval before initiating simultaneous interpretation can be shortened significantly, enabling earlier delivery of information to the listeners.

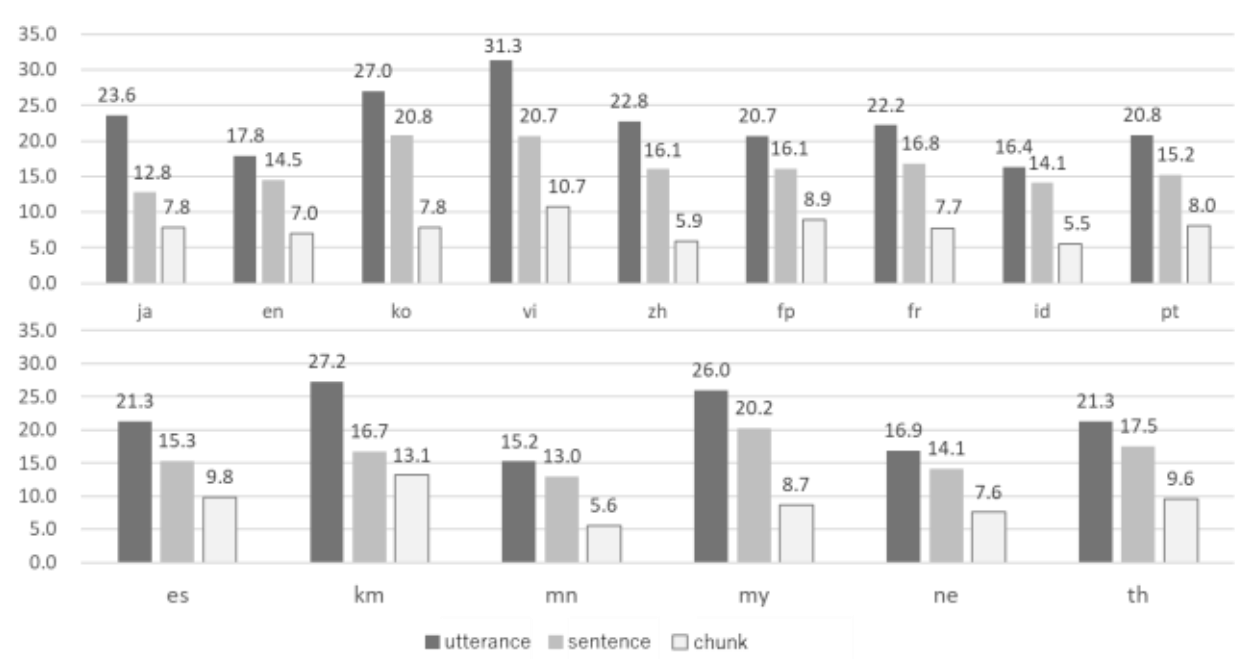

**Figure 4:** Average segment length in simultaneous interpretation test sets

In addition, we assessed the translation quality of the multilingual chunk segmentation outputs by feeding them into a MT engine and evaluating the results using BLEU scores (Papineni et al., 2002).

The test sets of the simultaneous interpretation corpus described in Section 2.2 were used for evaluation. Multiple utterances were concatenated, and the chunk translation results of the following four MT engines were evaluated using BLEU. The BASE engine is a baseline system that segments the input by a fixed length, i.e. 7 words, which is the average number of words per chunk in the test sets of the simultaneous interpretation corpus.

- TSEG-C: chunk segments and context tags
- TSEG-S: sentence segments and context tags
- GPMT: sentence segments, no context tags
- BASE: fixed-length (= 7 words) segments without context tags using GPMT engine

Figure 5 summarizes the evaluation results for the translation from Japanese into all other languages and from all other languages into Japanese. For all translation directions, the TSEG-S engine translating sentence segments with contextual processing achieved the highest BLEU score, significantly outperforming the BASE and GPMT engines.

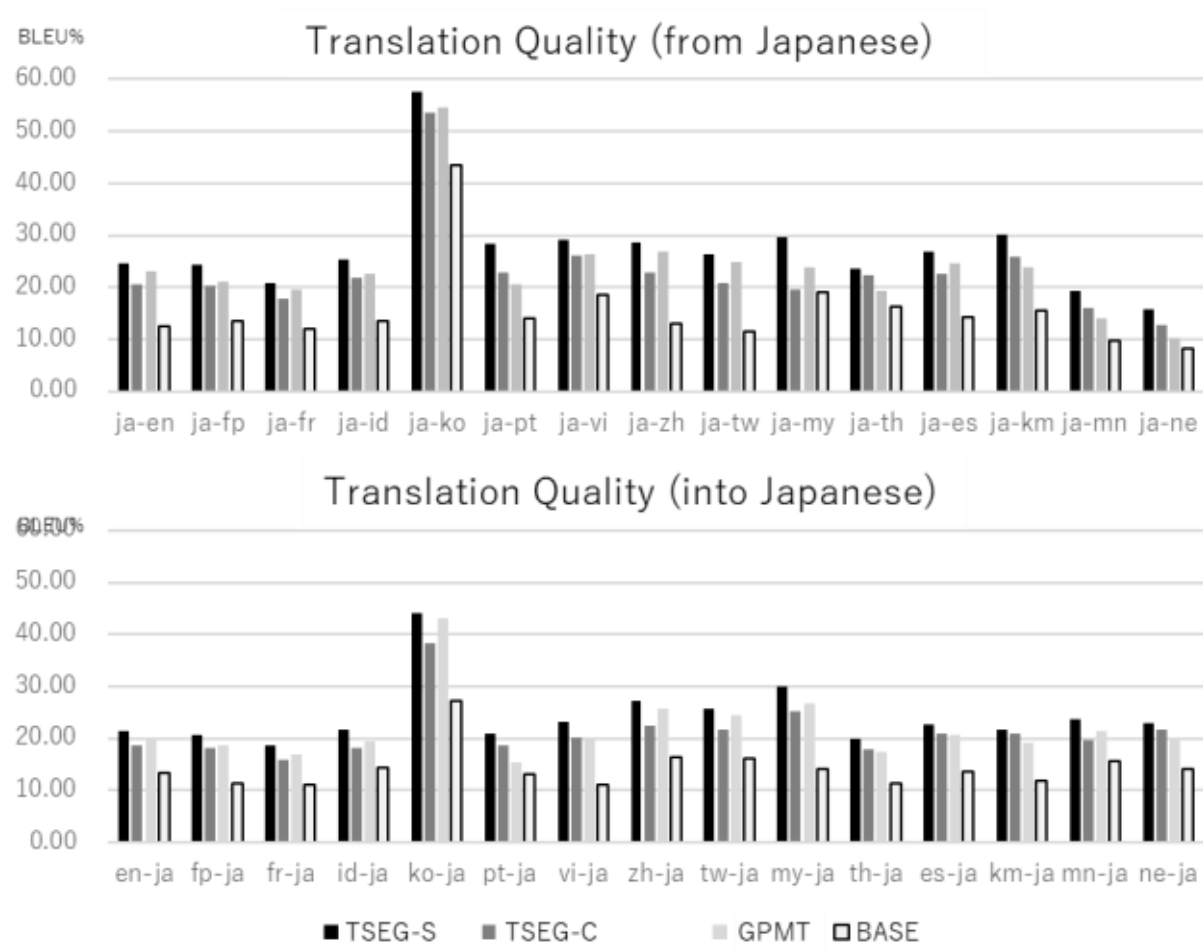

**Figure 5:** Automatic evaluation results

The contextual processing resulted in an average gain of 3.03 points compared to the GPMT engine. In particular, the quality of translations for several Asian languages (Khmer +6.78, Filipino +5.77, Nepali +5.49, Mongolian +5.22, etc.) has improved largely.

Regarding input segmentation, the TSEG-S engine was found to have an average advantage of 10.84 points over the fixed-length split (BASE) due to the usage of the online segmentation technique.

On the other hand, the translation results of the chunk-based TSEG-C engine obtained an average BLEU score of 3.68 points lower than TSEG-S. As described in Section 4.1, it is inevitable that the translation quality is lowered by chunk translation in exchange for the shortening of the response time.

However, the decline in translation accuracy of the TSEG-C engines is much smaller than the significant drop in performance when using the BASE engine. This suggests that by performing chunk-level segmentation at appropriate positions, TSEG-C effectively minimizes the degradation in MT quality.

In addition to the chunk-level translations, the system also presents high-quality retranslated outputs at the end of each sentence, thereby realizing a simultaneous interpretation system that delivers optimal and easily understandable interpretations for the listener.

### 4.3 Simultaneous Interpretation Services

The described core technologies have been employed by several private companies providing the following simultaneous interpretation services at Expo 2025 Osaka.

- "EXPO Honyaku" is a translation app used in one-to-one multilingual conversation scenes.
- "EXPO Honyaku Remote" is a one-to-many service that was used for guided tours hosted by the Expo Association.
- "EXPO Simultaneous Interpretation System" which simultaneously interprets the contents of presentations and displays subtitles in real time at World Exposition seminars.
- Multilingual chat communication at virtual Expo venues
- "Simultaneous Interpretation Experience Booth" with advanced UI/UX using avatar interaction.

These services were positioned within the “Digital Expo” initiative of the “Future Society Showcase Project” presenting a showcase of future society by incorporating advanced technologies and systems.

NICT will continue to advance these technologies and leverage the Expo achievements to accelerate technology transfer and social implementation.

## Acknowledgement

Part of this work was conducted as part of the research and development project “Advancement of Multilingual Translation Technologies” under the Ministry of Internal Affairs and Communications ICT Priority Technology R&D Project (JPMI00316).